%% file: main.tex
\documentclass[lettersize,journal]{IEEEtran}
\usepackage{amsmath,amsfonts}
\usepackage{algorithmic}
\usepackage{algorithm}
\usepackage{array}
\usepackage[caption=false,font=normalsize,labelfont=sf,textfont=sf]{subfig}
\usepackage{textcomp}
\usepackage{stfloats}
\usepackage{url}
\usepackage{authblk}
\usepackage{verbatim}
\usepackage[utf8]{inputenc}
\usepackage[pdftex]{graphicx}

\usepackage{graphicx}
\usepackage{cite}
\title{Enhancing multimodal analogical reasoning with Logic Augmented Generation}
\author[1]{Anna Sofia Lippolis}
\author[2]{Andrea Giovanni Nuzzolese}
\author[1]{Aldo Gangemi}

\affil[1]{University of Bologna, Italy}
\affil[2]{ISTC-CNR, Italy}
\begin{document}


\markboth{Journal of \LaTeX\ Class Files,~Vol.~14, No.~8, August~2021}%
{Shell \MakeLowercase{\textit{et al.}}: A Sample Article Using IEEEtran.cls for IEEE Journals}


\maketitle

\begin{abstract}
Recent advances in Large Language Models (LLMs) have demonstrated their capabilities across a variety of tasks. However, automatically extracting implicit knowledge from natural language remains a significant challenge, as machines lack direct experience with the physical world. Given this scenario, semantic knowledge graphs can serve as conceptual spaces that guide LLMs to achieve more efficient and explainable results. In this paper, we apply a Logic Augmented Generation (LAG) framework that leverages the explicit representation of a text through a semantic knowledge graph and applies it in combination with prompt heuristics to elicit implicit analogical connections. This method generates extended knowledge graph triples representing implicit meaning and enabling systems to reason on unlabeled multimodal data regardless of the domain. We validate our work through three metaphor detection and understanding tasks across four datasets, as they require deep analogical reasoning capabilities. The results show that this integrated approach surpasses current baselines and performs better than humans in understanding visual metaphors. It also enables to obtain justifications for the reasoning processes, though still has inherent limitations in metaphor understanding, especially for domain-specific metaphors. Furthermore, we propose a thorough error analysis, discussing issues with existing metaphor datasets, metaphorical annotations, and current evaluation methods.
\end{abstract}

\begin{IEEEkeywords}
Logic Augmented Generation, Figurative Language Understanding, Analogical Reasoning, Large Language Models, Conceptual Blending Theory
\end{IEEEkeywords}

\section{Introduction}
\label{sec:intro}
\input{sections/intro}

\section{Related works}
\label{sec:background}
\input{sections/background}


\section{Methodology}
\label{sec:methodology}
\input{sections/methodology}

\section{Application and Evaluation}
\label{sec:eval}
\input{sections/eval}

\section{Discussion}
\label{sec:discussion}
\input{sections/discussion}

\section{Conclusion}
\label{sec:conclusion}
\input{sections/conclusion}

\bibliographystyle{IEEEtran}
\bibliography{references}

\section{Biography Section}

\begin{IEEEbiographynophoto}{Anna Sofia Lippolis}
Anna Sofia Lippolis is a PhD student at the University of Bologna and ISTC-CNR in Bologna, 40136, Italy. Her research interests include semantic web applications, digital humanities, and knowledge engineering automation with Large Language Models. Contact her at annasofia.lippolis2@unibo.it.
\end{IEEEbiographynophoto}

\begin{IEEEbiographynophoto}{Andrea Giovanni Nuzzolese}
Use $\backslash${\tt{begin\{IEEEbiographynophoto\}}} and the author name as the argument followed by the biography text.[Full name] is [role] at [institution] at [city, state, postal code, country]. [His/Her] research interests include [3 very brief (not a complete list of) topics]. [Last name] received [his/her] [highest degree] in [topic] from [institution]. [He/She] is a [member/fellow/other] at [professional organization]. Contact [him/her] at [website or email address].
\end{IEEEbiographynophoto}

\begin{IEEEbiographynophoto}{Aldo Gangemi}
Use $\backslash${\tt{begin\{IEEEbiographynophoto\}}} and the author name as the argument followed by the biography text.[Full name] is [role] at [institution] at [city, state, postal code, country]. [His/Her] research interests include [3 very brief (not a complete list of) topics]. [Last name] received [his/her] [highest degree] in [topic] from [institution]. [He/She] is a [member/fellow/other] at [professional organization]. Contact [him/her] at [website or email address].
\end{IEEEbiographynophoto}

\end{document}

%% file: sections/intro.tex

%
From musical improvisation to solving complex scientific problems, much of what we know and do relies on implicit knowledge: the vast, unconscious reservoir of information we acquire through experience rather than formal instruction\footnote{Implicit and tacit knowledge are usually treated as synonyms. For a more comprehensive account on the distinction between the two see the work by Davies~\cite{davies2015knowledge}.}. Consequently, as philosopher Polanyi claimed, ``we know more than we can tell'' \cite{polanyi2009tacit}.
Analogical reasoning operates on this implicit knowledge base by identifying structural similarities between different domains and shaping how we interpret the world, draw inferences, identify conceptual links, and even understand figurative language~\cite{kaufman2010implicit}. Unlike humans, however, machines struggle with these tasks, as they lack direct experience of the physical world, making the computational modeling of analogical reasoning a significant challenge\footnote{As shown by the emergence of dedicated conference workshops on the matter, such as the Workshop on Analogical Abstraction in Cognition, Perception, and Language (Analogy-Angle): \url{https://analogy-angle.github.io/}}.
    
\begin{figure}[htbp]
    \centering
    \caption{Evaluation framework combining automated metrics, expert assessment, and human validation across four layers of analogical reasoning: detection, comprehension, reasoning, and generalization.}
    \includegraphics[width=0.5\textwidth]{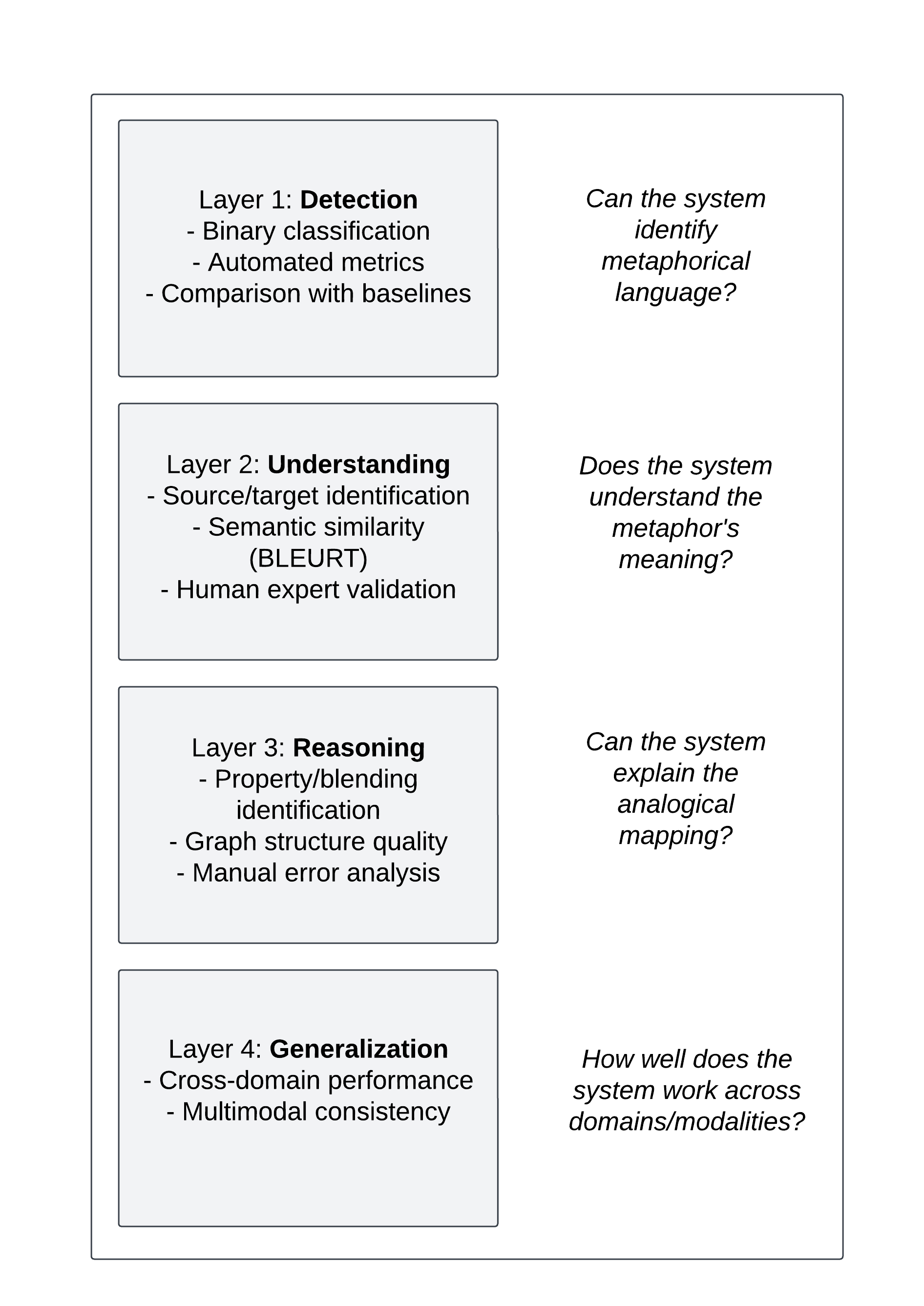}
    
    \label{fig:process}
\end{figure}
Metaphor is arguably the most demanding form of analogical reasoning because it requires not only structural mapping but also the creative blending of distinct conceptual domains \cite{wijesiriwardene2023analogical}. Large Language Models (LLMs) struggle with this task: current prompting methods capture surface‐level word statistics but miss the compositional and experiential grounding needed for genuine analogy \cite{yang2024elco,keysers2019measuring,furrer2020compositional}. Although these models are trained on vast, heterogeneous corpora that contain fragments of implicit meaning, analogical knowledge is not explicitly codified during training and therefore persists only as probabilistic associations in their embedding space—a dependence that ultimately constrains their reasoning \cite{MAO2025102712}. Their limited grasp of relational links is a known reasoning deficit that standard benchmarks rarely expose \cite{nezhurina2024alice}. In fact, recent narrative-consistency work shows that, despite stellar token-level scores, frontier LLMs collapse under long-range coherence checks; metaphor requires the same variety of global relational reasoning \cite{ahuja2025finding}. Therefore, despite repeated calls to enrich AI systems with deeper semantic and cognitive representations of the world, rather than mere word-level correlations \cite{schrimpf2021neural,lake2021word,rule2020child,leivada2023sentence}, analogical reasoning—especially in the form of metaphor—remains a major open challenge for Natural Language Processing \cite{stowe-etal-2021-metaphor,leivada2023sentence}. Progress is further hampered by the absence of a shared problem definition for metaphor and by the scarcity of diverse, well-curated datasets and evaluation protocols \cite{hicke2024science}.

Yet these challenges may be addressable through hybrid neurosymbolic approaches. 
Research shows that LLMs' performance improves significantly with structured background material~\cite{xie2023next,liu2024educating}. Consequently, combining reliable but labor-intensive semantic technologies with faster, context-dependent yet error-prone LLMs—as demonstrated by Logic Augmented Generation (LAG) in the work by Gangemi and Nuzzolese~\cite{GANGEMI2025100859}—opens a promising new avenue for metaphor understanding and analogical reasoning.


Building on Conceptual Blending Theory—an extension of Lakoff and Johnson’s Conceptual Metaphor Theory that models how multiple mental spaces fuse into novel meaning \cite{fauconnier2003conceptual}—we inject an explicit blending ontology into a LAG pipeline (Section \ref{sec:methodology}) to capture the semantic implicit relationships underlying metaphors from unlabelled multimodal data. Integrating structured, machine-interpretable semantics through KGs with fast and dynamic LLMs systems allows the real-time flexible understanding and representation of the implied aspects of metaphor in data across various domains and modalities such as news, visual advertisements, memes, and medical sciences. This hybrid architecture can be classified as a Type 2–3 neurosymbolic system in Kautz’ taxonomy of neurosymbolic systems~\cite{kautz2022third}.  


The contributions of this work are: (i) the adaptation of the LAG framework to enhance multimodal analogical reasoning; (ii) its multi-dimensional evaluation, as shown in Figure \ref{fig:process}, on variations on the task of metaphor detection and understanding, to which we include a novel dataset to test scientific metaphors; and (iii) a discussion on its computational implementation and applications\footnote{All supplementary material, code and data are available on GitHub at \url{https://github.com/dersuchendee/metaphorical-logic-augmented-generation}. Data is also archived on Zenodo with the following DOI: \url{10.5281/zenodo.15355852}}. Our findings show that while we achieve high performance in metaphor detection and generally outperform baselines, the inherent limitations of current LLMs unveil at deeper understanding decreases at higher reasoning levels, with particular challenges in domain-specific and visual metaphors.

The structure of the paper is as follows: Section \ref{sec:intro} introduces the significance and contributions of this work; Sections \ref{sec:background} and III provide the related works and an overview of the theoretical frameworks of analogical abstraction in metaphors; Section \ref{sec:methodology} describes the LAG methodology employed in the paper, Sections \ref{sec:eval} and VI concern its experimental application to metaphor detection tasks. Section \ref{sec:discussion} discusses the results, and Section \ref{sec:conclusion} summarizes the contributions and implications for extracting implicit meaning to reason analogically from figurative unlabelled data.

%% file: sections/background.tex
This section reviews prior work on conceptual blending, metaphor knowledge representation and automated metaphor processing, with a particular focus on detection and interpretation rather than generation.

\subsection{Foundational approaches to Computational Blending Theory}

Early computational approaches to CBT established theoretical foundations that inform current work. Goguen's algebraic semiotics provided one of the first rigorous mathematical models of conceptual blending, representing metaphors as combinations of structured sign systems and formally specifying how conceptual domains can be merged~\cite{goguen1998introduction}. Building on this foundation, Kutz and colleagues extended CBT into ontological frameworks, developing methods for algorithmically merging distinct domain ontologies to create novel hybrid concepts through shared structural constraints~\cite{confalonieri2020blending,kutz2012ontological}. Veale's computational models pioneered practical implementations of CBT, extending metaphor interpretation systems to implement conceptual integration with a focus on generative creativity~\cite{veale2001computation,veale2019conceptual}. While these approaches established crucial theoretical groundwork, they primarily focus on creative concept invention and generation rather than metaphor detection and understanding, with the latter being our primary concern.

\subsection{Knowledge representation of metaphor}

Early knowledge representation efforts handled metaphors in isolation, using SWRL rules to derive new facts from fixed triples \cite{Hamilton2021,Mitrovic2017}.  Newer lines of work align metaphors with frame resources: Panayiotou~\cite{pana} links cross-domain figures through bibliographic ontologies; MetaNet\footnote{\url{https://metaphor.icsi.berkeley.edu}} anchors conceptual metaphors to FrameNet frames; and Amnestic Forgery embeds MetaNet in Framester, yielding 642 mapped metaphors across 270 source and 221 target frames \cite{gangemi_18}. Despite this progress, static taxonomies are inherently incomplete and need to be complemented by the intervention of systems that can process metaphors dynamically.  

\subsection{Theory-integrated metaphor detection and understanding} 

Recent advancements in metaphor detection have increasingly incorporated theoretical frameworks into computational approaches. Studies leveraging Conceptual Metaphor Theory (CMT) have shown improved baseline results in metaphor identification~\cite{mao2023metapro,tian2024theory}. For metaphor interpretation, various methods have been developed to discern the source domain implied by the metaphor, given a target domain. These include unsupervised techniques \cite{shutova2017annotation}, deep learning-based approaches~\cite{rosen2018computationally}, and methods using Large Language Models \cite{wachowiak2023does}. However, these methods generally depend on annotations of a single domain rather than exploiting unlabeled data. In Hicke et al.,~\cite{hicke2024science}, when unlabeled data was exploited, the Metaphor Identification Procedure was preferred as a theoretical framework, thus the metaphors were annotated by category but not by source and target domain. Indeed, the authors recognize this limitation, stating that with the Metaphor Identification Procedure, forcing the models to annotate word-by-word makes it challenging for them to identify metaphors comprising multi-word units.
Other research focuses on extracting properties linking the source and target domains to interpret metaphors~\cite{su2017automatic,rai2019understanding,su2020deepmet}. While these approaches yield promising results, they often do not adequately capture the underlying theoretical reasoning processes that are based on the definition of their structural similarity in terms of abstraction and relational dependency on the two metaphorical domains. Additionally, for these systems to work, they tend to require a specified target word.
In the context of visual metaphor understanding, while the work by He et al.~\cite{he2024viemf} focuses on detection, it does not adequately address metaphor understanding according to existing theoretical frameworks, which claim that a comprehensive approach should at least involve identifying both source and target domains.
To foster richer research in visual metaphors, the MetaCLUE task set was introduced~\cite{akula2023metaclue}, though its code and dataset are not publicly available. Metaphoric compositions linked by an annotated attribute are available in the ELCo Dataset proposed by Yang et al.~\cite{yang2024elco}, where metaphorical composition in emojis is evaluated via a textual entailment task. 
Integrating heuristics based on existing ontologies on metaphor and knowledge graphs into the metaphor detection and understanding pipeline remains largely unexplored, despite its potential to render theories standardized and machine-readable across multimodal contexts, reduce hallucinations, adapt them for other automated reasoning tasks, and enhance system explainability.

\subsection{Hybrid approaches for computational metaphor processing}

In recent years, researchers have grafted structured semantic knowledge onto Transformer backbones to improve metaphor detection and interpretation.  
FrameBERT \cite{li2023framebert} fine-tunes \texttt{RoBERTa} while jointly learning embeddings for FrameNet lexical units, yielding interpretable token–frame alignments.  
Lieto et al. introduce MET\textsuperscript{CL}—a T\textsuperscript{CL}-based system that builds a single, emergent prototype for each metaphor by composing \textsc{head} and \textsc{modifier} concepts \cite{lieto_delta_2025}. The approach is anchored in the FrameNet-style frames embedded in MetaNet and therefore inherits the blind spots of that inventory: if no source/target frame exists,  MET\textsuperscript{CL} must first extend MetaNet with heuristic candidates before it can even run its logic. MET\textsuperscript{CL}’s prototypes are assembled from ConceptNet triples, and risk importing properties that are irrelevant to the discourse at hand. 


\section{Theoretical foundations}
In this section, we explore the background of our model and discuss the challenges related to existing automated processes for analogical abstraction in the context of metaphor.

\subsection{Implicit knowledge and analogical abstraction}

Implicit knowledge is a type of knowledge acquired independently of conscious attempts to do so and that goes beyond the ability to explain it \cite{reber1989implicit,polanyi2009tacit}. 

Analogical abstraction is a kind of implicitly learned knowledge where patterns or analogies are detected across experiences or concepts. This recognition of structural similarities in the process of analogical reasoning can be used to solve problems and make accurate decisions about novel circumstances, and is particularly clear when the connections between concepts are not obvious or are learned through repeated exposure.

A fundamental component of NLP for many years, LLM-based analogical reasoning has been recently investigated. Webb et al. show that LLMs excel at identifying simple proportional word analogies~\cite{webb2023emergent}. However, other studies show LLMs' performance diminishes when faced with more complex rhetorical figures~\cite{wijesiriwardene2023analogical}, such as metaphors, or when the task requires distinguishing between associative and relational responses~\cite{stevenson2023large}. These findings suggest that while LLMs can identify associations learned from data, they do not necessarily grasp the underlying relationships that define how concepts are connected in more abstract ways. As a result, there is a need for a structured framework that directs these systems to undertake specific analytical steps during tasks that require analogical abstraction while making space for explainability.

\subsection{Conceptual Blending Theory, metaphor, and frame semantics}
According to Conceptual Metaphor Theory~\cite{lakoff1981}, metaphors create parallels between domains, highlighting experiential correlations.
This claim, along with Mental Spaces Theory, is expanded by CBT, which describes one of the basic cognitive operations of human beings: conceptual combination. According to CBT, in metaphor, a generic space makes abstract and links the two source and target input spaces and generates a new blended space according to a metaphorical criterion or ``property"~\cite{fauconnier2003conceptual}.
Indeed, many studies~\cite{forceville2002pictorial,glucksberg2008metaphors,Black1993} claim that by explaining one object in terms of another, the ‘target’ domain receives a property from the other object, the ‘source’, or in general that the source and target are associated in thought and language because they share a property of some kind.  For example, in the metaphor \textsc{ideas are food}, ideas and food are connected through the criterion of \textit{internalization} (food is inserted and processed in our bodies like ideas), and the two spaces are abstracted with respect to the agent, the input, the process, the output, selectively basing on the context.
In this way, the connecting property (\textit{internalization}) is a higher-level concept that transcends the inherent characteristics of each input space and motivates the comparison.



However, the existence of a shared property is by no means a sufficient condition for two concepts to participate in a metaphorical mapping~\cite{dancygier2016figurativeness}: for metaphorical meaning to bridge between individuals and within a community, those individuals must already share some understanding of the components involved in the metaphor~\cite{colston2023roots} due to commonsense knowledge and/or specific cultural or social context in which a metaphor is used~\cite{petridis2019human}. In this regard, the CBT framework can be explicitly connected to Fillmore's frame semantics~\cite{fillmore_06}, where certain verbs automatically invoke particular roles, values, and perspectives of discourse participants~\cite{avelar2020metaphors}. Here, different communicative contexts activate background knowledge (i.e. frames) that constrains interpretation. This frame-shifting mechanism central to metaphorical interpretation is therefore well-explained by these theoretical frameworks. 

\subsection{Logic Augmented Generation}
The so far described limitations of theory-integrated metaphor detection and understanding studies can be addressed with the integration of ontologies and commonsense KGs in an LLM-based metaphor detection pipeline. The LAG approach, as shown in Gangemi and Nuzzolese~\cite{GANGEMI2025100859}, conceptualises LLMs as potential Reactive Continuous Knowledge Graphs (RCKGs), which can dynamically adapt to diverse inputs by extending and contextualising Semantic Knowledge graphs (SKGs) that work as base models. LAG exploits the duality between implicit knowledge  generated during training and explicit knowledge, externalised in the form of structured knowledge graphs through generation and prompting. 
RCKG extraction involves a three-step process, i.e. (i) mapping multimodal signal to natural language, (ii) converting natural language into an SKG, and (iii) extending the SKG with tacit knowledge based on multiple heuristics (the extended knowledge graph as XKG).

In this frame of reference, SKGs ensure logical consistency, enforce factual boundaries, and foster interoperability, while LLMs process unstructured data and provide contextual insights on implicit knowledge on demand.

In summary, our LAG approach represents a Type 2-3 neurosymbolic system \cite{kautz2022third} that dynamically generates domain-specific knowledge graphs and ontology-based heuristics to guide LLM reasoning, rather than relying on static knowledge bases or simple knowledge retrieval like other hybrid approaches. Unlike purely symbolic approaches that lack flexibility in handling novel metaphors, or purely neural approaches that lack explainability in their reasoning processes, LAG bridges this gap by using structured knowledge representations to guide neural generation while maintaining interpretability. The integration of CBT with SKGs enables our framework to handle metaphor processing task across both textual and visual modalities, addressing the limitations of existing approaches that typically focus on single tasks or modalities. Furthermore, by generating explicit knowledge graphs that represent the blending process, our approach provides explanatory capabilities that current LLM-based methods lack, making the analogical reasoning process transparent and verifiable.


\


%% file: sections/methodology.tex
The proposed method, based on LAG \cite{GANGEMI2025100859}, aims to enhance the detection and understanding of metaphor by establishing a standardized and explainable model for the semantic representation of implicit natural language. In this way, the tacit \textit{know-how} knowledge, following Polanyi's definitions, can become \textit{know-that} \cite{polanyi2009tacit}. Our approach has two main components: i) the representation of inputs, regardless of their modality, into natural language, ii) their conversion into a SKG with Text2AMR2FRED and iii) the SKG enhancement driven by CBT applied to metaphor.
\vspace{-1em}
\begin{figure}[h] 
  \centering 
  \includegraphics[width=0.5\textwidth]{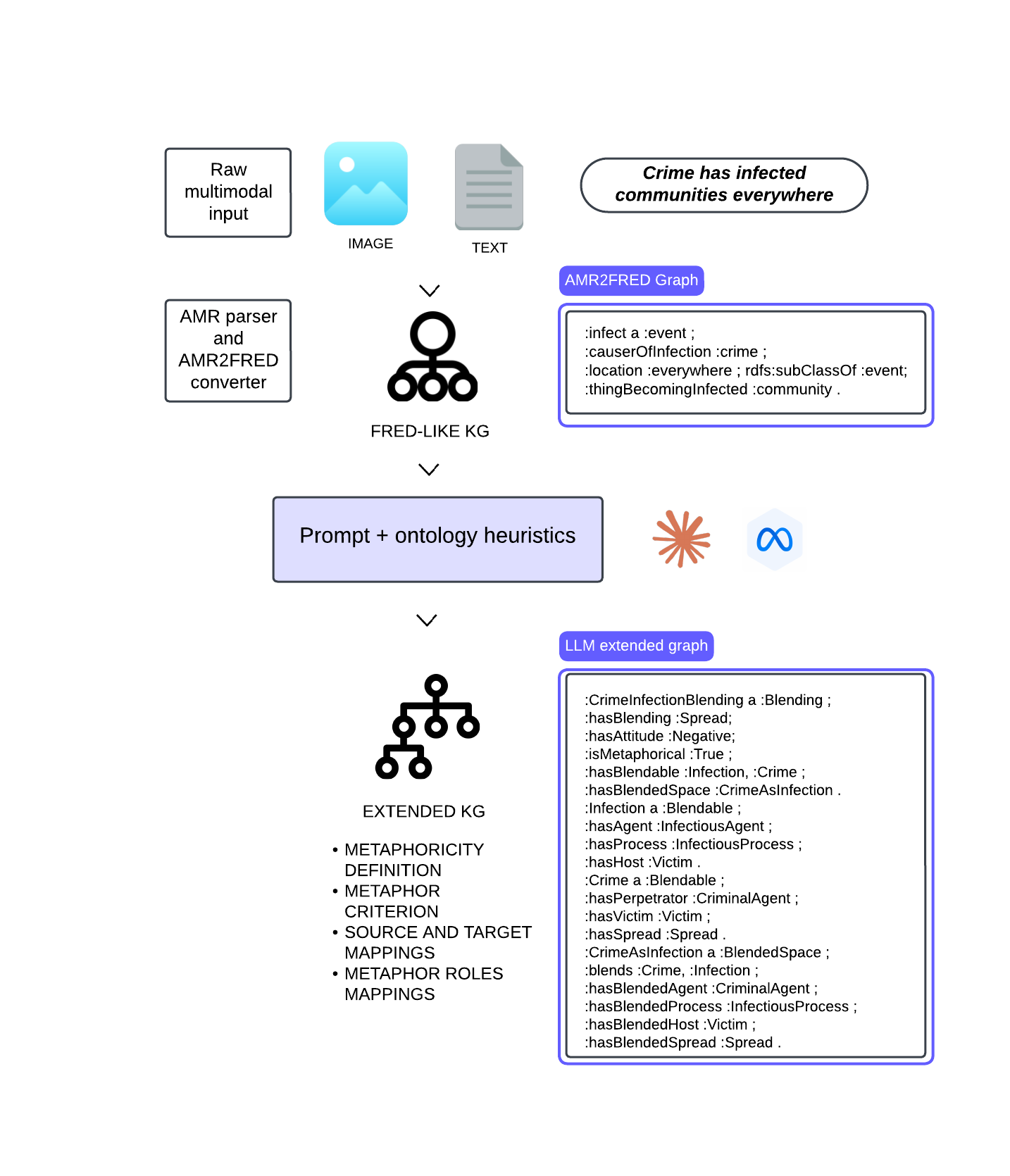} 
  \caption{Proposed LAG-based pipeline for the sentence ``Crime has infected communities everywhere''.} 
  \label{fig:pipeline} 
\end{figure}


\subsection{Automated multimodal representation with Text2AMR2FRED}

Text2AMR2FRED has been first introduced as a revised architecture of FRED's text-to-KG pipeline~ \cite{gangemi2023text2amr2fred}, using state-of-the-art Abstract Meaning Representation (AMR) parser with multilingual capabilities. In this way, relevant commonsense knowledge from knowledge bases linked by the Framester knowledge base \cite{gangemi2016framester} is retrieved and represented in a graph.

For example, as shown in Fig. \ref{fig:pipeline}, the sentence ``Crime has infected communities everywhere'' is first parsed into an AMR graph, using SPRING~ \cite{blloshmi2021spring}. The AMR graph is then converted into an RDF/OWL knowledge graph that follows FRED's knowledge representation patterns using AMR2FRED. 

AMR identifies the cause of the infection in the sentence, which is, however, not a germ, but crime and the thing that gets infected (community) everywhere. In the FRED-like KG, this graph is enriched with knowledge from Framester such as frames, WordNet synsets and PropBank roles, along with alignment to the DOLCE foundational ontology~\cite{gangemi2002sweetenin}. However, there is no representation of metaphorical knowledge yet at this stage. While a conceptual metaphor KG sourced from Metanet is contained in Framester, it is still necessary to be able to detect figurative language in any type of data.
An image can also be used by being automatically described in natural language. In this case, language is an amodal way to bridge modalities in the resulting supramodal SKG. The SKG can then be extended with implicit contextual knowledge with the support of an LLM.

\subsection{Logic Augmented Generation with heuristics}

As shown in Fig. \ref{fig:pipeline}, the result of the framework that generates XKG triples is the extension of the SKG with new classes and properties according to a specific ontology in the prompt functioning as heuristics. In this work, we adopt the Blending Ontology for this purpose.

\subsubsection{The Blending Ontology}
The Blending Ontology\footnote{Available at \url{https://github.com/dersuchendee/BlendingOntology}} is based upon mDnS (mereological Descriptions and Situations)\footnote{\url{http://www.ontologydesignpatterns.org/ont/mdns/mdns.owl}}, the Cognitive Perspectivisation ontology as defined in Gangemi and Presutti~\cite{vossen_fokkens_2022}\footnote{\url{http://www.ontologydesignpatterns.org/ont/persp/perspectivisation.owl}}, and the MetaNet schema from the Amnestic Forgery ontology by Gangemi et al.~\cite{gangemi2018amnestic}. 
By introducing the concepts of descriptions and situations, the Description and Situations ontology design pattern~\cite{gangemi2003understanding} provides a formalization of frame semantics where a situation is a set of facts as described by an observer that is interpreted by a description, a schema that defines the entities observed in a situation.
Situations can be reported from different perspectives: the Cognitive Perspectivisation Ontology proposed by Gangemi and Presutti~\cite{vossen_fokkens_2022} represents the \texttt{Perspectivisation} frame, a type of situations where a fact is reported within a certain narrative, or \texttt{Lens}, which creates a viewpoint, cutting the reality which a \texttt{Conceptualiser} holds an attitude towards. Perspectivisation itself is therefore not merely rhetorical, as it blends entities playing two roles: the \texttt{cut} and the \texttt{lens}.
Finally, the Amnestic Forgery ontology~ \cite{gangemi2018amnestic} is a model for metaphor semantics based on MetaNet and compatible with Conceptual Metaphor Theory \cite{lakoff1981metaphors}. Amnestic Forgery reuses and extends the Framester schema, allowing to deal with both semiotic and referential aspects of frames.





The Blending Ontology defines key classes and properties that operationalize the blending of concepts:

\begin{itemize}
  \item \textbf{Classes} such as \texttt{Blend}, \texttt{Blendable}, \texttt{Blended}, and \texttt{Blending} represent different stages and elements of the conceptual blending process. 
  
  \item \textbf{Object Properties} such as \texttt{blendableComponent},
  
  \texttt{blendedComponent}, and \texttt{blendingComponent}  are used to link these classes in meaningful ways that reflect the underlying cognitive processes. These properties help in mapping the components of one conceptual domain to another, leading to the emergence of new, blended concepts. Other properties like \texttt{enablesBlending} and \texttt{inheritsRoleFrom} define the interaction between different conceptual elements within the ontology.
\end{itemize}

\begin{figure}[htbp]
    \centering
    \includegraphics[width=0.48\textwidth]{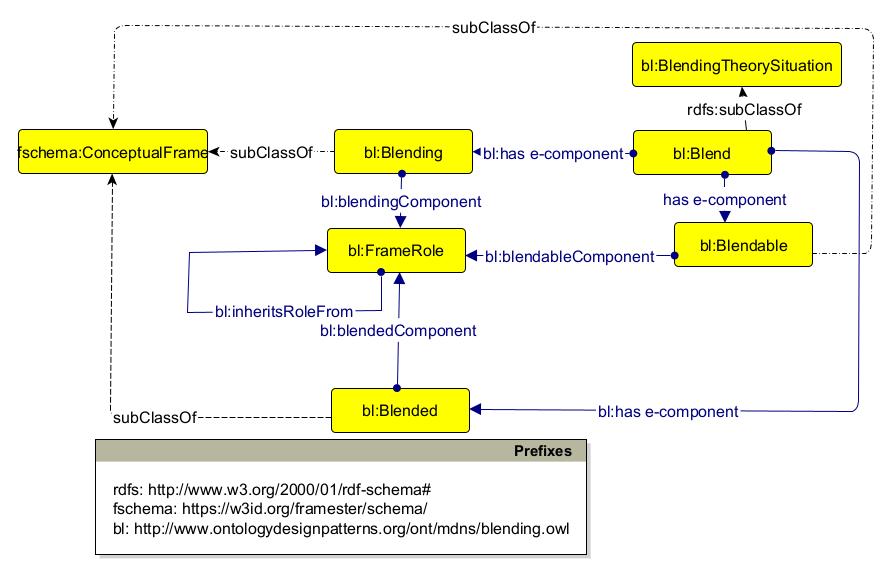} 
    \caption{The core of the Blending ontology\protect\footnotemark.}
\label{fig:blending-ontology}
    \label{fig:blending}
\end{figure}

More specifically, as shown in Fig. \ref{fig:blending}, the basic Descriptions include:
\begin{enumerate}
    \item \texttt{Blending}, a generic description enabling Blendable frames to be mapped by subsuming some of their components;
    \item \texttt{Blendable}, two descriptions with components that are subsumed by components in the Blending frame;
    \item \texttt{Blended}, the description that inherits components from the \texttt{Blendable} frames, firstly through the \texttt{Blending} sharable space, and eventually inheriting other components from the \texttt{Blendable} frames, giving rise to a novel set of blended situations. As a consequence, the \texttt{Blended} is considered as a novel entity, separately from the \texttt{Blendable};
    \item \texttt{Blend}, the meta-description that has as components \texttt{Blending}, \texttt{Blendable}, and \texttt{Blended}.
\end{enumerate}

\footnotetext{The full diagram can be seen on Github \url{https://github.com/dersuchendee/metaphorical-logic-augmented-generation}}
When a conceptual metaphor is formulated from two input frames, their roles are generalized and inherited (a \texttt{Blendable} \texttt{inheritsRole}) from a more generic \texttt{Blending}, which enables the blending (\texttt{EnablesBlending}). The \texttt{Blended} in turn inherits components from the \texttt{Blendable} frames through the \texttt{Blending} generic space. The \texttt{Blended} is therefore a new, independent space where the metaphor takes place.

Such fine-grained element mappings are required to understand the consistency of the blend, something that has been mentioned in research but so far has yet to be implemented~\cite{wachowiak2023does}. The \texttt{Blending} frame motivating the metaphorical blend also makes a system's analogical reasoning capabilities more explainable. For example, a system can correctly identify the elements involved in a metaphor but misidentify the property that links them, as discussed in Section \ref{sec:discussion}, showing the inability to understand the metaphor. With this framework in mind, it is possible to create a prompt for an LLM that, given a text and the resulting KG representation from Text2AMR2FRED, it extends it to create an augmented RDF graph in Turtle format. 



%% file: sections/eval.tex
In this section, we evaluate the LAG-based framework’s ability to extract implicit meaning by applying it to metaphor detection and understanding tasks. We first use three metaphor detection datasets to determine the best-performing configuration among several baselines. This optimal configuration is then applied to three metaphor understanding datasets. 

\subsection{Tasks}

This section describes the application of the framework in three metaphor-related tasks: i) \textit{metaphor detection}, where a SKG derived from a sentence in natural language is enhanced to determine if a sentence is metaphorical, sometimes with a target word as a reference; ii) \textit{conceptual metaphor understanding}, which additionally aims to identify the correct source and target domains that make up the metaphor, and iii) \textit{visual metaphor understanding}, where the aim is to identify the property linking the two domains in the visual metaphor. 


\subsubsection{Metaphor detection and understanding}

As shown in the Supplementary Material, in this task, the LLM is given a sentence, a SKG representing the sentence derived from Text2AMR2FRED, and a series of instructions.
The prompt first requires identifying the possible source and target of the sentence; then, it is asked to define these two as conceptual frames containing roles and following a blending function that orchestrates the creation of a blended space. Following the Blending Ontology's alignment to the Cognitive Perspectivization one (See Section \ref{sec:methodology}), a \texttt{Lens} and an \texttt{Attitude} further refine the definition of the metaphor. An example of the output graph for a specific sentence is also provided.

Given the Text2AMR2FRED graph, and the presumed source and target of the sentence given a target word when available, along with blending-derived instructions, the LLM is prompted to analyze the sentence and to use at least these specific ontology elements according to the given definitions, to determine the possible metaphoricity of a sentence: \texttt{bl:Blending}, \texttt{bl:Blendable}, \texttt{bl:Blended}, \texttt{cp:Attitude}, \texttt{cp:Lens}, and the datatype property \texttt{metanet:isMetaphorical} which the classification is based on. In the prompt, we also ask to determine blendable roles and their mapping to create a blended metaphorical space according to the Blending Ontology.

\subsubsection{Visual metaphor understanding}
\begin{figure}[h!]
  \centering
  \caption{Example prompts and extracts of correct and wrong outputs according to manual evaluation for visual metaphor understanding.} 
  \includegraphics[width=0.5\textwidth]{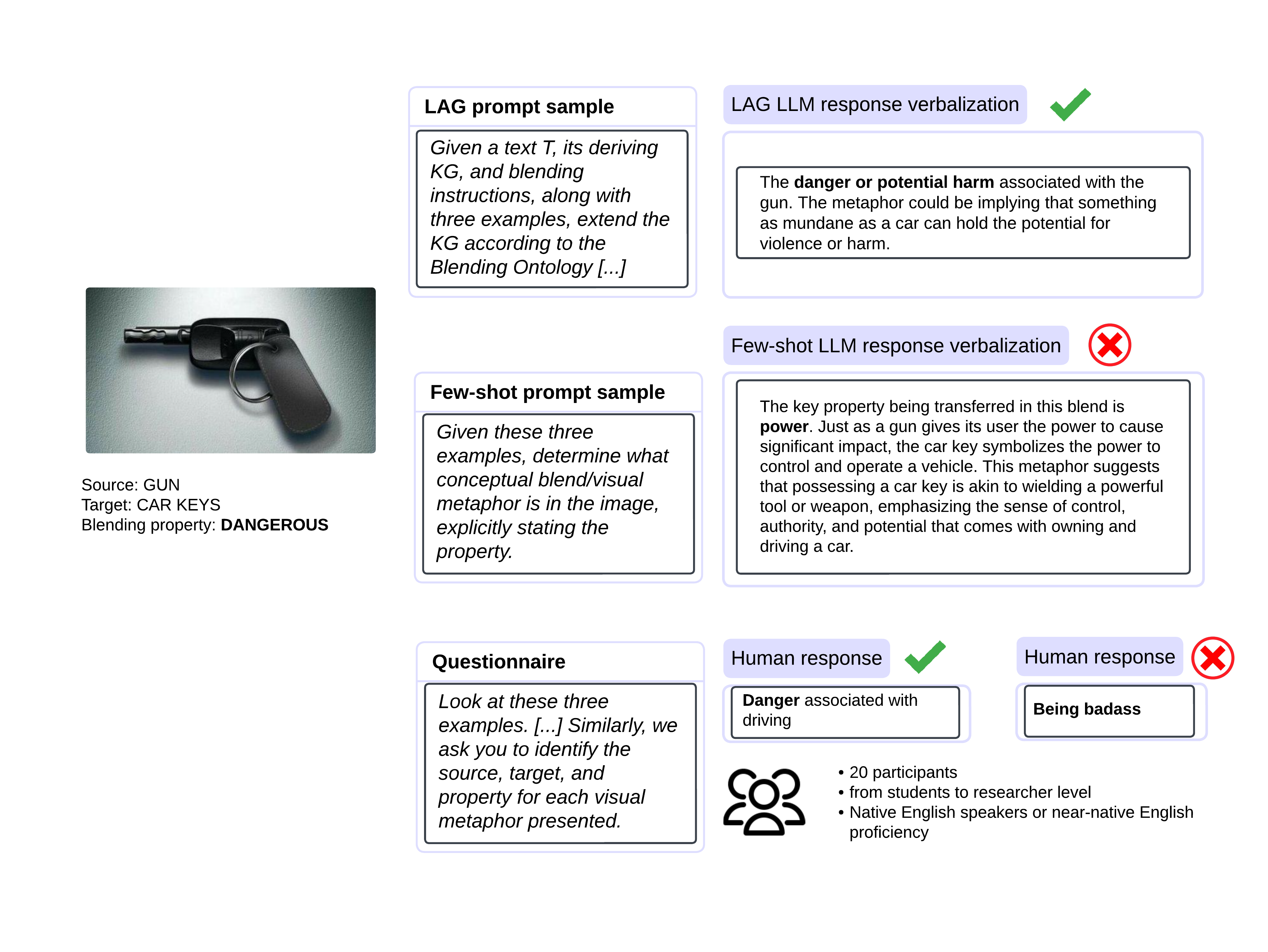} 
  \label{fig:visualmet2} 
\end{figure}

Following the original study performed on human participants outlined in Petridis et al.~\cite{petridis2019human}, the methodology for visual metaphor understanding involves the presentation of three examples and correct annotations of visual metaphor elements within an in-context learning technique, along with a new image to analyze.
In this case, each image to analyze is first described in natural language through the LLM, and the text is passed to Text2AMR2FRED to generate a SKG. The SKG, image, accompanying sentence, and Blending Ontology heuristics are all included in the prompt with the three examples to generate the XKG, as shown in Fig. \ref{fig:visualmet2}. In this work, we experiment with multiple combinations of these inputs to find the best prompting technique.
\vspace{-1em}
\section{Datasets}

For the purpose of this study, and for comparison across the outlined tasks, for textual metaphor detection we conducted experiments on two publicly available metaphor datasets: (1) MOH-X~\cite{mohammad2016metaphor}, which comprises 647 sentences annotated as either metaphorical or literal; and (2) TroFi~\cite{birke2006clustering}, another metaphor detection dataset collected from the 1987-1989 Wall Street Journal Corpus. We sampled 300 balanced instances each for testing.
For conceptual metaphors, we employ (3) the dataset refined by Wachowiak and Gromann~\cite{wachowiak2023does} of 447 sentences including conceptual metaphors annotated with source and target domain (for ease of reference, we will name it \textit{WG}).
We also introduced a fourth dataset to verify the capability to identify and explain domain-specific conceptual metaphors: (4) the Balanced Conceptual Metaphor Testing Dataset (BCMTD). This dataset consists of 147 sentences, equally distributed among general conceptual metaphors (derived from annotated Metanet-derived data from Framester~\cite{gangemi2016framester} and that weren't present in WG), scientific conceptual metaphors (extracted from scientific literature from the Appendix in Van Rijn and Van Tongeren~\cite{vanrijn1997metaphors} and examples from Semino et al.~\cite{semino2018metaphor}, allowing a domain-specific dataset on medical metaphors), and literal sentences from the VUA corpus~\cite{steen2010method} (previously refined by Wachowiak and Gromann~\cite{wachowiak2023does} as many of the sentences were wrongly labeled or deemed unfit for the task). The visual metaphors dataset is derived from Petridis et al.~\cite{petridis2019human}. It contains 48 images, 27 ads and 21 non-ads, and three examples, that have been given to human participants to test their ability to identify visual metaphors according to a conceptual blending framework. 
Table \ref{tab:dataset_summary} shows statistics for all the datasets.

\begin{table}[h]
\centering
\caption{Summary of datasets statistics, including instances, percentage of metaphorical sentences, and samples.}
\label{tab:dataset_summary}
\resizebox{0.95\columnwidth}{!}{%
\begin{tabular}{|p{3cm}|p{2cm}|p{2cm}|p{2cm}|}
\hline
\textbf{Dataset} & \textbf{\# Instances} & \textbf{\% Met.} & \textbf{\# Samples} \\ \hline
MOH-X & 647 & 48.7 & 300 \\ \hline
TroFi & 3737 & 43.5 & 300 \\ \hline
WG & 447 & 100 & 447 \\ \hline
BCMTD & \begin{tabular}[c]{@{}c@{}}170 (conceptual)\\ 48 (scientific)\\ 50 (VUA)\end{tabular} & 66.6 & 147 \\ \hline 
Visual metaphors & 51 & 100 & 48 \\ \hline 
\end{tabular}%
}
\end{table}



\subsection{Evaluation methodology}

The evaluation methodology for metaphor detection is both automated and manual depending on the type of task. 

\subsubsection{Automated evaluation metrics}
We use Accuracy and F1 score to evaluate the model's performance in metaphor detection. For conceptual metaphor understanding, evaluated via the correct identification of both source and target mappings according to a gold standard, the BLEURT score~\cite{sellam2020bleurt} is used in order to evaluate the semantic distance between the candidate and reference source and target in the metaphor and considered to be successful if both achieve a positive score. 
This is an evaluation method similar to the one in Saakyan et al.~\cite{saakyan2022report}, where instead of only reporting label accuracy, it is reported according to defined thresholds. We also compute the correlation between the BLEURT score and human evaluation. 





\subsubsection{Manual evaluation}
To correctly determine metaphorical source and target mappings for domain identification, and to correctly compare gold standard and predicted blending properties for visual metaphor interpretation, it is also necessary to conduct a manual evaluation. The best performing solution for conceptual and visual metaphor understanding is therefore evaluated for source and target mappings by three annotators. The annotators were graduated in linguistics, had proven expertise on metaphors, and had English Native or near-native language proficiency. They exhibited a Fair inter-annotator agreement (Fleiss' Kappa) in all cases, with an average on all tasks of 0.35, which is understandable for a task so subjective as metaphor understanding. To select the gold standard label, we employ majority voting between the annotators. More information on the agreement scores is in the Supplementary Materials. To calculate the Accuracy, we compute an average of each Accuracy score. 
In the visual metaphor understanding task, the manual evaluation compares the gold standard property with the predicted one connecting source and target. The evaluation directly follows the one in Petridis et al.~\cite{petridis2019human}, which considered as correct responses directly related to the property in the image. 
We also repeated the study on a sample of 24 images on 20 participants of the University of Bologna, manually evaluating their responses as well. Participants ranged from student to researcher level, with native or near-native English fluency. This setup slightly differs from the original study, where participants were employed without specifying their study level. Participants' submissions were anonymous and informed consent was provided.
Furthermore, we provide a sample of 16 generated graphs, 8 derived from correctly identified metaphorical sentences, 8 derived from visual metaphors, to two knowledge engineers experts in conceptual blending and conceptual metaphor theories for an evaluation of the quality of the graphs.
We evaluate the representation’s adequacy, correctness, and completeness on a Likert scale from 1 to 7 concerning detected property, source, target and element mappings. Finally, one knowledge engineer manually evaluated errors (derived from negative majority voting) for conceptual metaphor understanding of the best performing LAG configuration. The errors follow the categorization proposed in Wachowiak and Gromann~\cite{wachowiak2023does} with the addition of a new category found, the \textit{incorrect switching of source and target} (for a comprehensive account of these categories, see the Supplementary Material).


\subsection{Baseline methods}


In Tian et al.~\cite{tian2024theory} the best-performing method for metaphor detection is a Theory Guided Scaffolding Instruction Framework based on Conceptual Metaphor Theory (TSI CMT). MetaPRO is a baseline among non-LLM methods for detecting source and target domains, serving as the state-of-the-art approach for identifying mappings without specifying a target word~\cite{mao2023metapro}. For conceptual metaphor understanding, no current baselines simultaneously detect source and target but only one of the two. We thus employ LLMs in different configurations, including zero-shot and few-shot, shown to be effective in previous research~\cite{wachowiak2023does}. No established baselines for visual metaphor understanding specify source, target, and blending properties. Thus, we explore different configurations of the LAG framework using the same three few-shot examples used to test humans in~\cite{petridis2019human}, and evaluate against the few-shot baseline without LAG.




\subsection{Parameters setup}

In our experiments, Claude 3.5 Sonnet\footnote{\url{https://www.anthropic.com/news/claude-3-5-sonnet}} serves as the primary LLM. We compare the best-performing results with Llama 3.1 70B-Instruct\footnote{\url{https://ai.meta.com/blog/meta-llama-3-1/}} in all cases except for visual metaphor interpretation, where no current open models support few-shot image examples. Only the APIs are used for testing. We conducted tests using default parameters and set the temperature to 0 for greedy decoding.

\subsection{Main results}
In this section, we outline the main results of the LAG framework applied to the selected tasks.

\subsubsection{Metaphor detection: comparison with baselines}

Table \ref{tab:performance_comparison} shows comparative results of LAG compared with baselines for the sampled TroFi and MOH datasets. Table \ref{tab:performance_bctmd} shows comparative results of LAG with respect to the BCMTD.

\begin{table}[ht]
\centering
\caption{Performance comparison of various methods for metaphor detection on MOH-X and TroFi. Best performing results are in bold.
}
\begin{tabular}{|l|cc|cc|}
\hline
\textbf{Method} & \multicolumn{2}{c|}{\textbf{MOH-X}} & \multicolumn{2}{c|}{\textbf{TroFi}} \\
\cline{2-5}
 & \textbf{F1 (\%)} & \textbf{Acc. (\%)} & \textbf{F1 (\%)} & \textbf{Acc.(\%)} \\
\hline
MetaPRO  & 84  & 81  & 79  &  70 \\
TSI CMT* & 82.5  & 82.9  & 66  & 66.8  \\
LAG & \textbf{89.7} & \textbf{87.3}  & \textbf{89.7}  & \textbf{84.6}  \\
\hline
\end{tabular}
\label{tab:performance_comparison}
\end{table}

\begin{table}[!h]
\caption{Performance metrics for the BCTMD Dataset for metaphor detection. Best performing results are in bold.}
\centering
\begin{tabular}{|c|c|c|}
\hline
\textbf{Method} & \textbf{ Accuracy (\%)} & \textbf{ F1 Score (\%)} \\ \hline
LAG                 & \textbf{80.1}               & \textbf{84.1}               \\ \hline
MetaPRO                & 69.1               &   69.8\\ \hline
Few-Shot 12                 & 59.0                & 48.9                \\ \hline
Few-Shot 6                 & 52.4                & 45.2               \\ \hline
Few-Shot 3                 & 47.5                & 42.8               \\ \hline
Zero-shot                 & 22.9               & 33.8               \\ \hline

\end{tabular}
\label{tab:performance_bctmd}
\end{table}


\subsubsection{Conceptual metaphor understanding}

Conceptual metaphor understanding is a more challenging task than metaphor detection, as it involves the correct identification of both source and target domains for a metaphor, which shows whether the system has understood it or not. 
\\
\textbf{WG dataset.}
Manual evaluation reveals that the percentage of metaphors where source and target were correctly identified at the same time is 25.6\%.
BLEURT scores showed a moderate correlation between the human annotations and the predicted domains.
For the target domain, a point-biserial correlation of 0.702 (\textit{p} $<$ 0.001) and a Spearman correlation of 0.698 indicate that BLEURT is reasonably sensitive to the human judgments.
For the source domain, a point-biserial correlation of 0.707 (\textit{p} $<$ 0.001) and a Spearman correlation of 0.680 were observed.

\textbf{BCTMD.}
Manual evaluation on the best performing method in metaphor understanding reveals that out of 96 sentences containing conceptual metaphors, the percentage of correctly identified source and target at the same time is 34.3\%.
Furthermore, the distribution of correctly identified metaphors is 51.6\% for generic conceptual metaphors, while for scientific conceptual metaphors the score lowers to 8\%. BLEURT scores showed a 36.9\% positive correlation between reference and predicted source and target domains. A point-biserial correlation of 0.3525 (\textit{p} $<$ 0.001) and Spearman correlation of 0.2917 (\textit{p} $<$ 0.001) both indicate moderate positive correlations for BLEURT.







\subsubsection{Visual metaphor understanding}

For visual metaphor understanding, the identification of the blending property corresponds to the correct interpretation of the source and target employed in the metaphor (as shown by Petridis et al.\cite{petridis2019human}). 
Out of the three configurations, shown in Table \ref{tab:method_accuracy}, manual annotation from the three evaluators revealed that the best performing result is LAG without the insertion of a sentence describing the image alongside the image itself in the prompt.
Concerning human evaluation described in Petridis et al.~\cite{petridis2019human}, which correctly interpreted the meaning of the image 41.32\% of the time, in the best performing method the LLM correctly interpreted the meaning an average of 67.06\% of the time. We then sampled the dataset, randomly taking 24 visual metaphors out of the 48 in the dataset, and repeated the original study ~\cite{petridis2019human} with 20 human participants. They have correctly interpreted the metaphorical property 59.25\% of the time, with respect to the average LLM score of 73.5\% for the same images.

\begin{table}[h]
\centering
\begin{minipage}{.45\linewidth}
\centering
\caption{Comparison of methods for visual metaphor understanding. The best performing result is LAG without the injection of a sentence that describes the image in the input.}
\begin{tabular}{|p{2cm}|p{1cm}|}
\hline
\textbf{Method} & \textbf{Accuracy (\%)} \\ \hline
LAG sent+img    & 65          \\ \hline
LAG no sent    & \textbf{67}          \\ \hline
LAG no img        & 65.2          \\ \hline
Few-Shot (3)     & 54.7   \\ \hline
\end{tabular}
\label{tab:method_accuracy}
\end{minipage}%
\hspace{0.05\linewidth}
\begin{minipage}{.45\linewidth}
\centering
\caption{Ablation study results for the visual metaphor dataset. Best results in bold.}
\begin{tabular}{|c|c|c|}
\hline
\textbf{Method} & \textbf{Acc.\%}  \\ \hline
LAG no sent & 67   \\ \hline
No Blending & \textbf{68.6}   \\ \hline
No Graph & 56.2  \\ \hline
\end{tabular}
\label{tab:ablation_compact2}
\end{minipage}
\end{table}


\subsubsection{Metaphor detection and understanding: comparison with open-source LLM}
In Table \ref{tab:llamacomp}, we compare the Claude-based best performing results of metaphor detection with the ones from Llama, 3.1 70B Instruct, showing Claude outperforms the open-source alternative.

\begin{table}[h]
\centering
\caption{Comparison of LAG for metaphor detection in each dataset with Claude and with Llama. Best results in bold.}
\resizebox{\columnwidth}{!}{ 
\begin{tabular}{|c|c|c|c|c|c|c|}
\hline
\textbf{LLM} & \multicolumn{2}{c|}{\textbf{MOH-X}} & \multicolumn{2}{c|}{\textbf{TroFi}} & \multicolumn{2}{c|}{\textbf{BCMTD}} \\ \cline{2-7} 
 & \textbf{Acc.\%} & \textbf{F1\%} & \textbf{Acc.\%} & \textbf{F1\%} & \textbf{Acc.\%} & \textbf{F1\%} \\ \hline
\textbf{Claude} & \textbf{87.3} & \textbf{89.7} & \textbf{84.6} & \textbf{89.7} & \textbf{80.1} & \textbf{84.1} \\ \hline
\textbf{Llama} & 55.9 & 69 & 60.8 & 75 & 66.6 & 66.6 \\ \hline
\end{tabular}
}
\label{tab:llamacomp}
\end{table} 
For what concerns conceptual metaphor understanding, according to the evaluators, the average percentage of Accuracy is 37.5\% against the method with Claude, which yields an Accuracy of 34.3\%.
The results show a drop in performance when using Llama with metaphor detection, but metaphor understanding of both source and target mappings seems to perform slightly better with respect to Claude.

\subsubsection{Ablation studies}

We conduct the ablation study to evaluate the impact of the components in our method for metaphor detection by two configurations: (i) removing the blending instructions, and ii) removing the knowledge graph injection in the prompts. All the prompts used in these configurations are in the Supplementary Materials. Table \ref{tab:ablation_compact} shows the results of the ablation studies for metaphor detection for all the textual datasets, while Table \ref{tab:ablation_compact2} for visual metaphor understanding.

\begin{table}[h]
\centering
\caption{Ablation study results for MOH-X, TroFi, and BCMTD datasets. Best results in bold.}
\resizebox{\columnwidth}{!}{ 
\begin{tabular}{|c|c|c|c|c|c|c|}
\hline
\textbf{Method} & \multicolumn{2}{c|}{\textbf{MOH-X}} & \multicolumn{2}{c|}{\textbf{TroFi}} & \multicolumn{2}{c|}{\textbf{BCMTD}} \\ \cline{2-7} 
 & \textbf{Acc.\%} & \textbf{F1\%} & \textbf{Acc.\%} & \textbf{F1\%} & \textbf{Acc.\%} & \textbf{F1\%} \\ \hline
\textbf{LAG} & \textbf{87.3} & \textbf{89.7} & \textbf{84.6} & \textbf{89.7} & \textbf{80.1} & 84.1 \\ \hline
\textbf{No Blending} & 81.6 & 87 & 81.9 & 86 & 78.6 & \textbf{85.2} \\ \hline
\textbf{No Graph} & 78.6 & 82 & 83.9 & 87 & 70 &  73\\ \hline
\end{tabular}
}
\label{tab:ablation_compact}
\end{table} 

\subsubsection{Quality of the augmented graphs and error analysis}\label{sec:errors}

The results of the survey provided to two knowledge engineers on the quality of the augmented graphs highlight a consistent pattern where textual components generally outperform their visual counterparts across all evaluated metrics. Specifically, on a Likert scale from 1 to 7, correctness and adequacy in text scored at 5.1 and 6.3, suggesting a positive assessment, whereas the visual aspects scored slightly lower at 4.9 and 5 respectively, indicating room for improvement. The completeness of the text was rated at 5.2, while the visual completeness was lower at 4.4, showing a need for better integration or representation of visual elements to match the textual analysis. 

For what concerns the error analysis, misalignment in conceptual metaphor mapping is predominantly due to \textit{wrong subelement mapping}, which accounts for approximately 56.5\% in the WG dataset and 57.1\% in the BCTMD. Annotations that are too specific contribute significantly as well, with 23.6\% and 28.6\% of errors in WG and BCTMD respectively, while overly general descriptions appear in 9.3\% of WG and 14.3\% of BCTMD cases. Less frequent errors concern switching the source and target (3.24\% in WG) and identifying as literal something that was a metaphor (6.94\% in WG). For visual metaphor understanding, the error analysis still follows the one done with human participants in Petridis et al.~\cite{petridis2019human}, with errors mostly involving \textit{incorrect objects} (21\% of the errors) and \textit{incorrect property} (57\% of the errors). In 21\% of cases, even when the LLM identifies source, target and connecting property, the role mapping explanation can show a misunderstanding of the metaphor (\textit{incorrect target symbol}). For a more detailed analysis along with example errors, see the Supplementary Material.

%% file: sections/discussion.tex
In this section, we discuss the results, outline some applications and further research directions for the proposed framework, and investigate limitations.

\subsection{Application of LAG to multimodal metaphor detection and interpretation}
The results outlined in Section \ref{sec:eval} show that for the task of metaphor detection, the proposed LAG-based framework outperforms the current metaphor detection and understanding baselines. 
This approach can be used to extract knowledge graphs from unstructured multimodal metaphorical data. As such, it could have effects in different applications, for example, improving hate speech, which often has metaphorical features~\cite{lemmens-etal-2021-improving}, and analyzing debates in the press and online media.
For what concerns visual metaphors, since humans tend to struggle to identify them~\cite{petridis2019human}, the automated understanding of visual blends can bridge this gap. 
Additionally, using an open-source LLM like Llama can decrease the performance concerning detection, but not understanding tasks. 

Ablation studies show that while removing the graph reduces the performance in all cases, removing the blending heuristics in the prompt reduces it in all cases besides visual metaphor understanding. 
While this is a slight difference, it could be motivated by the fact that visual metaphors can be wrongly described in text before processing. 
Another explanation for this slight performance drop is shown by the error analysis (see Section \ref{sec:errors}), where \textit{incorrect property} was the most frequent issue both for humans~\cite{petridis2019human} and LLMs: the objects are detected in the images, but there is still a struggle with extracting the correct properties. Because half of the images are advertisements, which typically present clearer context, the intended properties of the objects are usually clearly defined. By not specifying the purpose of the image and its context, we make it challenging for both humans and machines to accurately determine the metaphorical property via a blending mechanism that tends to be more open-ended without contextual specifications.
Additionally, the low accuracy score observed in detecting scientific metaphors appears to reflect inherent limitations in current LLMs rather than a specific weakness of the LAG approach. Scientific metaphors often require domain-specific context and specialized knowledge that may not be well-represented in LLM training data. Also, scientific metaphors show multiple levels of conventionality, tending to result more literal as they come to be widespread in language.

\subsection{Quality of the augmented graphs and error analysis}\label{sec:errors}
The results from the questionnaire to analyze the quality of the augmented graphs suggest a positive performance with a disparity between text and visual outputs, probably because the chosen examples were the ones taken by Petridis et al.~\cite{petridis2019human} and they weren't further annotated with an RDF syntax in order to be consistent with the human evaluation.

For what concerns error analysis, overly general descriptions show difficulty in identifying the correct taxonomical level of the metaphor. Additionally, less frequent errors, such as switching the source and target and identifying as literal something that was a metaphor, show further layers of complexity in the metaphor mapping process, such as misunderstanding the role of source and target domains and determining the metaphoricity level (a sentence might seem literal when the perceived conventionality of the metaphor is high). 


These results underscore the need to train and test automatic methods on domain-specific figurative datasets, but also that analogical reasoning is generally not properly performed in these tasks. On the other hand, visual metaphor understanding shows different kinds of errors. Errors involving \textit{incorrect objects} indicate poor image understanding skills, for example, a LLM will identify the shape of a gun as a whistle, or pink Nike shoes as fishes, usually skewing the interpretation. The most common error, however, is \textit{incorrect property}: if the LLM correctly identiﬁes the objects, source and target, but still misinterprets the property. For example, in the case of the blending of the car keys and the gun (displayed in Figure \ref{fig:visualmet2}), with the annotated property being \textsc{dangerous}, the predicted property is \textsc{powerful}. This error is also observed in human participants and tied to their values and beliefs over the objects displayed in the blend, a further area of investigation. Furthermore, the presence of the \textit{incorrect target symbol} error shows that the evaluation of the property alone as proposed in previous studies~\cite{petridis2019human} is not sufficient to verify if the system has understood a metaphor.

The error analysis shows the need to incorporate contextual meaning into the understanding of the metaphor, but also indicates that complete metaphor understanding requires a deeper level of semantic and cultural awareness that despite guidance towards relational analogical thinking, current systems struggle to achieve, also due to the absence of standardized gold-standard annotations that would establish clear metaphorical hierarchies. This multifaceted challenge highlights the gap between computational pattern recognition and the cognitive processes underlying human metaphor comprehension.

\subsection{Interpretability of metaphorical datasets}
The proposed framework's explainability through XKG creation helps to outline some aspects that so far haven't fully emerged in computational approaches to metaphor.

\subsubsection{Context and culture}
Our research demonstrates that annotating the source and target elements in computational metaphor analysis is insufficient for fully grasping metaphorical meaning. For instance, as illustrated in Figure \ref{fig:visualmet2}, the LLM accurately identifies all objects within the image that form the metaphor. However, although the blending property is annotated as \textsc{dangerous}, the LLM interprets it as \textsc{powerful}. In the absence of external context that can further frame the reference elements needed for comprehension, both interpretations are arguably valid.

This and other examples from our experiments reveal two key insights. First of all, the perspective adopted by a LLM in interpreting a metaphor reflects its underlying cultural background. 
Furthermore, for systems (and users) to reliably converge on metaphorical annotated properties, providing additional context is essential. This requirement for context may also help explain the relatively poor performance observed in human responses, such as distinguishing whether an advertisement is a public campaign, a comic strip, or a standard advertisement.
    
\subsubsection{The dynamics of metaphor annotation}
Beyond culture and context-dependence, the human annotation showed that some sentences' gold standard metaphors in the WG, BCTMD and the Visual Metaphors dataset had multiple possible correct answers or even multiple answers in the same sentence. For example, for the sentence ``This mini-controversy erupted when Republicans introduced a string of amendments in a final effort to obstruct passage of the reconciliation bill'' the gold-standard metaphor is \textsc{inhibiting enactment of legislation is impeding motion}, while the predicted one is \textsc{controversy is volcano eruption}, which has been noted as correct by the evaluators but would be deemed as incorrect with automatic assessment. This phenomenon shows the need to keep both the human and the metaphor interpretation process flexible. However, current annotated metaphorical datasets only contain one possible gold standard answer. Future work should address the possibility to incorporate multiple correct answers for a reference text.  

\subsection{Future application of LAG to metaphor generation}
In addition to detecting existing metaphors, our framework offers a structured approach to generating new effective metaphors. By formalizing the process of blending conceptual frames, the system can suggest novel metaphors by creatively reconfiguring known concepts into new, plausible configurations according to a specified ontology. This aspect of the framework holds particular promise for applications in creative AI fields, such as advertising, literature, and automated content generation. In future research, we aim to develop a method for generation that addresses limitations found in other studies, such as the ones described in previous studies~\cite{chakrabarty2023spy,lieto_delta_2025}. 
Specifically, the authors in Lieto et al.~\cite{lieto_delta_2025} note that top-down frame taxonomie must be complemented by more flexible generators.
Chakrabarty et al. prompt models to generate visual metaphors through generic, implicit elements, but do not specify how these should be effectively combined to produce meaningful metaphors. Indeed, one reported error in the study was the omission of the target domain or essential elements that were not explicitly requested in the output image. Finding the LLM's pitfalls in understanding visual metaphors can thus improve its ability to generate metaphors in a closer way to our thought processes. This enrichment can foster narrative creation, offering fresh and impactful ways to communicate complex ideas or emotions, and enhancing the engagement and relatability of AI-generated content. 

\subsection{Limitations}
Despite its strengths, the current framework has possible improvements. Experiments are limited to English; more multimodal data is needed for a more comprehensive assessment. While the work on conceptual metaphors shows some correlation with automated measures, manual evaluation remains essential. Indeed, another survey on the quality of the augmented graphs with a higher quantity of graphs and expert participants is needed. Additionally, the model's computational complexity and resource demands may cause time constraints with large datasets or real-time applications. Ongoing refinement is thus necessary to improve scalability and efficiency.
For what concerns data leakage, while legacy textual benchmarks (e.g., MOH-X, TroFi) may already reside in the analyzed LLMs' web-scale pre-training corpus, the BTCMD corpus and the visual-metaphor dataset were newly curated and remained unpublished online before evaluation, ensuring no train–test leakage for these two resources.

%% file: sections/conclusion.tex
In this paper, we introduced a Logic-Augmented Generation pipeline that turns multimodal data into extended knowledge graphs by injecting a Blending Ontology at prompt time. Tested on four metaphor benchmarks, the system outperformed baselines by up to +12 $F_1$ on text and +8 accuracy on visual metaphors, and even surpassed the human gold standard on the visual metaphors dataset. Performance on scientific metaphors, however, trailed generic tasks by 17 percentage points, showing that domain-specific metaphors need to be treated differently from generic ones.

The error analysis supports recent findings that today’s LLMs excel at surface associations but falter on relational reasoning; many failures stem from sparse or mismatched contextual cues in existing corpora. Despite this shortcoming, which calls for more diverse and fine-grained metaphorical data, leveraging the Blending Ontology to generate augmented knowledge graphs enhances output explainability compared to current methods, which often fail to clarify why, even when the objects of a metaphor are detected, the metaphorical property connecting the two is not correctly identified.
The proposed approach advances computational models for multimodal figurative language understanding in unstructured data, with significant implications for AI in natural language processing, creativity, and applications such as hate speech detection. 